\documentclass[letterpaper]{article} % DO NOT CHANGE THIS
\usepackage{aaai23}  % DO NOT CHANGE THIS
\usepackage{times}  % DO NOT CHANGE THIS
\usepackage{helvet}  % DO NOT CHANGE THIS
\usepackage{courier}  % DO NOT CHANGE THIS
\usepackage[hyphens]{url}  % DO NOT CHANGE THIS
\usepackage{graphicx} % DO NOT CHANGE THIS
\usepackage{natbib}  % DO NOT CHANGE THIS AND DO NOT ADD ANY OPTIONS TO IT
\usepackage{caption} % DO NOT CHANGE THIS AND DO NOT ADD ANY OPTIONS TO IT
\usepackage{bm}
\usepackage{amssymb}
\usepackage{enumitem}
\usepackage{amsmath}
\usepackage{subfigure}
\usepackage{multirow}
\usepackage{appendix}
\usepackage{enumitem}
\usepackage[ruled, vlined, boxed, linesnumbered]{algorithm2e}
\usepackage{bibentry}
\usepackage{newfloat}
\usepackage{listings}
\DeclareCaptionStyle{ruled}{labelfont=normalfont,labelsep=colon,strut=off} % DO NOT CHANGE THIS
\title{Effective and Stable Role-Based Multi-Agent Collaboration by Structural Information Principles}
\author{
    Xianghua Zeng\textsuperscript{\rm 1},
    Hao Peng\textsuperscript{\rm 1},
    Angsheng Li\textsuperscript{\rm 1, 2}
}
\affiliations{
    \textsuperscript{\rm 1} State Key Laboratory of Software Development Environment, Beihang University, Beijing, China;\\
    \textsuperscript{\rm 2} Zhongguancun Laboratory, Beijing, China.\\
    \{zengxianghua, penghao, angsheng\}@buaa.edu.cn, 
liangsheng@gmail.zgclab.edu.cn.
}

\begin{document}

\maketitle

\begin{abstract}
Role-based learning is a promising approach to improving the performance of Multi-Agent Reinforcement Learning (MARL).
Nevertheless, without manual assistance, current role-based methods cannot guarantee stably discovering a set of roles to effectively decompose a complex task, as they assume either a predefined role structure or practical experience for selecting hyperparameters.
In this article, we propose a mathematical \textbf{S}tructural \textbf{I}nformation principles-based \textbf{R}ole \textbf{D}iscovery method, namely \textbf{SIRD}, and then present a SIRD optimizing \textbf{MARL} framework, namely \textbf{SR-MARL}, for multi-agent collaboration.
The SIRD transforms role discovery into a hierarchical action space clustering.
Specifically, the SIRD consists of structuralization, sparsification, and optimization modules, where an optimal encoding tree is generated to perform abstracting to discover roles.
The SIRD is agnostic to specific MARL algorithms and flexibly integrated with various value function factorization approaches.
Empirical evaluations on the StarCraft II micromanagement benchmark demonstrate that, compared with state-of-the-art MARL algorithms, the SR-MARL framework improves the average test win rate by $0.17\%$, $6.08\%$, and $3.24\%$, and reduces the deviation by $16.67\%$, $30.80\%$, and $66.30\%$, under easy, hard, and super hard scenarios.
\end{abstract}

\section{Introduction}\label{Introduction}
Cooperative multi-agent reinforcement learning (MARL) has broadly been applied to a variety of complex decisions, such as games \cite{nowe2012game,AlphaStar2019}, sensor networks \cite{zhang2011coordinated}, social network \cite{peng2022reinforced}, emergent tool usage \cite{baker2019emergent}, etc.
% , traffic management \cite{singh2020hierarchical}
There are two significant challenges in cooperative MARL: scalability that the joint state-action space grows exponentially with the number of agents \cite{yang2020overview}, and partial observability, which necessitates decentralized policies based on local action-observation histories because of communication constraints \cite{nguyen2020deep}.
Hence, the paradigm of Centralized Training with Decentralized Execution (CTDE) \cite{oliehoek2008optimal, kraemer2016multi, mahajan2019maven} is designed to deal with the above challenges.
However, the centralized training requires searching agent policies in the joint state-action space, making the CTDE challenging to learn efficiently, especially when the number of agents is large \cite{samvelyan2019starcraft}.
A valid solution is integrating roles to decompose tasks in multi-agent systems, where each role is associated with a particular subtask and a role policy in the restricted state-action space \cite{deloach2010mase, cossentino2014handbook, sun2020reinforcement}.

The key is bringing forward a set of roles to decompose the cooperative task effectively.
The typical methods to predefine task decomposition or roles \cite{lhaksmana2018role, sun2020reinforcement} require prior knowledge (subtask-specific rewards or role responsibilities) that is unavailable in practice.
It is impractical to automatically learn an appropriate set of roles from scratch, equivalent to learning in the joint state-action space with substantial explorations \cite{wilson2010bayesian, wang2020roma}.
Instead of learning roles from scratch, RODE \cite{wang2020rode} proposed to use the clustering technology to achieve role discovery in the joint action space.
However, the performance of the RODE depends heavily on practical experience due to its high sensitivity to parameters of adopted clustering algorithms, such as the cluster number of K-means, the maximum and minimum cluster number of X-means, and the neighborhood radius and density threshold of DBSCAN \footnote{Although recently proposed methods \cite{karami2014choosing, zhang2022automating}, achieving automatical parameters search, introduces significant computational overhead, which is not conducive to real-time decision-making.}.
Because practical task decomposition is not always well-defined a priori or changing, current role-based methods \cite{sun2020reinforcement, wang2020roma, wang2020rode} cannot guarantee effective role discovery without manual assistance.

This paper proposes a mathematical \textbf{S}tructural \textbf{I}nformation principles-based \textbf{R}ole \textbf{D}iscovery method, namely \textbf{SIRD}, and then presents a SIRD optimizing \textbf{MARL} framework, namely \textbf{SR-MARL}, for multi-agent collaboration.
The crucial insight is that, instead of flat clustering, the SIRD transforms role discovery into hierarchical action space clustering.
We firstly construct a weighted, undirected, and complete action graph, whose vertices represent actions, edges represent action correlations, and edge weight quantifies the functional correlation between actions with respect to achieving team targets.
Secondly, we leverage the one-dimensional structural entropy minimization principle \cite{li2016three} to sparsify the action graph, to generate an initial encoding tree.
Thirdly, we minimize the $K$-dimensional structural entropy \cite{li2018decoding} of the sparse graph to get the optimal encoding tree, thereby achieving the hierarchical action space clustering.
Furthermore, we take the hierarchical clustering on the optimal encoding tree as hierarchical abstracting of actions for role discovery.
The SIRD is independent of manual assistance and flexibly integrated with various value function factorization approaches.
Extensive experiments are conducted on the StarCraft II micromanagement tasks, including five easy maps, four hard maps, and four super hard maps.
Comparative results and analysis demonstrate the performance advantages of our proposed role discovery method.
All source code and data are available at Github\footnote{\url{https://github.com/RingBDStack/SR-MARL}}.
% publicly

The main contributions of this paper are as follows:
1) To the best of our knowledge, it is the first time to incorporate the mathematical structural information principles into MARL to improve effectiveness and stability under cooperative scenarios.
2) An innovative method, which transforms role discovery into a hierarchical abstracting on the encoding tree guided by one-dimensional and $K$-dimensional structural entropy, is proposed to optimize MARL for collaborative decisions.
3) Compared with the existing action space clustering, the hierarchical clustering achieved by the encoding tree guarantees more effective role discovery without manual assistance.
4) The extraordinary performance on challenging tasks shows that the SR-MARL achieves up to $6.08\%$ average test win rate improvement and up to $66.30\%$ deviation reduction than state-of-the-art baselines.

\section{Preliminaries}\label{PRELIMINARIES}

\subsection{Dec-POMDP}
In this work, we consider a fully cooperative multi-agent task $M$, which can be modeled as a Dec-POMDP \cite{oliehoek2016concise}.
The Dec-POMDP is defined as a tuple $\langle N, S, A, P, \Omega, O, R, \gamma\rangle$, where $N \equiv\{n_{1},n_{2}, \ldots, n_{|N|}\}$ is the finite set of agents, $S$ is the finite set of global states, $A$ is the action space, $P$ is the global state transition function, $\Omega$ is the finite set of partial observations, $O$ is the observation probability function, $R$ is the joint reward function, and $\gamma \in[0,1)$ is the discount factor. 
At each timestep, each agent $n_{i} \in N$ chooses an action $a_{i} \in A$ on a global state $s \in S$, and all chosen actions form a joint action $\bm{a} \equiv\left[a_{i}\right]_{i=1}^{|N|}$.
The joint action $\bm{a}$ results in a joint reward $r=R(s, \bm{a})$ and a transition to the next global state $s^{\prime} \sim P(\cdot \mid s, \bm{a})$.
We consider the partially observable setting, where each agent $n_{i}$ receives an individual partial observation $o_{i} \in \Omega$ according to the function $O\left(o_{i} \mid s, a_{i}\right)$. 
Each agent $n_{i}$ has an action-observation history $\tau_{i} \in \mathcal{T}$ and constructs individual policy $\pi_{i}\left(a_{i} \mid \tau_{i}\right)$ to maximize team performance jointly.
% The formal objective is to find a joint policy $\bm{\pi}\equiv\left[\pi_{i}\right]_{i=1}^{|N|}$ that maximizes the joint action-value function $Q^{\boldsymbol{\pi}}(s, \boldsymbol{a})$.

\subsection{Role-based Learning}
The idea of role-based learning is to decompose a multi-agent cooperative task into a set of subtasks, where the role set specifies task decomposition and subtask policies \cite{wilson2010bayesian, wang2020rode}.
Given a fully cooperative multi-agent task $M$, let $\Psi$ be the role set.
Each role $\rho_{j} \in \Psi$ is defined as a tuple $\left\langle t_{j}, \pi_{\rho_{j}}\right\rangle$, where $t_{j}$ is a subtask defined as $\langle N_{j}, S, A_{j}, P, \Omega, O, R, \gamma\rangle$, $N_{j} \subset N, \cup_{j} N_{j}=N$, and $N_{j} \cap N_{l}=\varnothing, j \neq l$.
$A_{j}$ is the restricted action space of role $\rho_{j}$, $A_{j} \subset A$, $\cup_{j} A_{j}=A$, and $|A_{j} \cap A_{l}| \geq 0$, $j \neq l$.
$\pi_{\rho_{j}}: \mathcal{T} \times A_{j} \mapsto [0,1]$ is the role policy for the subtask $t_{j}$.
% The role-based MARL aims to find $\Psi$ to maximize the expected global return $Q^{\Psi}\left(s, \boldsymbol{a}\right)$.

\subsection{Structural Information Principles}
In the structural information principles \cite{li2016structural}, the structural entropy measures the uncertainty of a graph under a strategy of hierarchical partitioning.
And the dynamics measurement of the encoding tree is recently analyzed \cite{Yang2022DynamicMO}.
By minimizing the $K$-dimensional structural entropy, the optimal hierarchical structure of the graph is generated, namely the optimal partitioning tree, which we call an ``encoding tree" in our article.
Given a weighted undirected graph $G=(V, E, W)$, $V$ is the vertex\footnote{A vertex is defined in the graph and a node in the tree.} set, $E$ is the edge set, and $W: E \mapsto \mathbb{R}^{+}$ is the weight function.
Let $n=|V|$ be the number of vertices and $m=|E|$ be the number of edges.
For each vertex $v \in V$, its degree $d_{v}$ is defined as the sum of the weights of its connected edges.
Then, we formally give the definitions of the encoding tree, the one-dimensional, and $K$-dimensional structural entropy of the weighted undirected graph $G$.

\subsubsection{Encoding Tree.}
The encoding tree of $G$ is defined as a rooted tree $T$ with the following properties:
1) For each node $\alpha \in T$, there is a vertex subset in $G$ corresponding with $\alpha$, $T_{\alpha} \subseteq V$.
2) For the root node $\lambda$, we set $T_{\lambda}=V$.
3) For each node $\alpha \in T$, its children nodes are marked as $\alpha^{\wedge}\langle i\rangle$ ordered from left to right as $i$ increases, and ${\alpha^{\wedge}\langle i\rangle}^{-}=\alpha$.
4) For each node $\alpha \in T$, we suppose that $L$ is the number of its children nodes; then all vertex subsets $T_{\alpha^{\wedge}\langle i\rangle}$ are disjointed, and $T_{\alpha}=\bigcup_{i=1}^{L} T_{\alpha^{\wedge}\langle i\rangle}$.
5) For each leaf node $\nu$, $T_{\nu}$ is a singleton containing a single vertex in $V$.

\subsubsection{One-dimensional Structural Entropy.}
The one-dimensional structural entropy of $G$ is defined as follows:
\begin{equation}\label{1d_se}
    H^{1}(G)=-\sum_{v \in V} \frac{d_{v}}{vol(G)} \cdot \log _{2} \frac{d_{v}}{vol(G)}\text{,}
\end{equation}
where $vol(G)=\sum_{v \in V} d_{v}$ is the volume of $G$.
% , the sum of degrees of all vertices in $V$

\subsubsection{$K$-dimensional Structural Entropy.}
Given an encoding tree $T$ whose height is at most $K$, the $K$-dimensional structural entropy of $G$ is defined as follows:
\begin{equation}\label{kd_se}
    H^{K}(G)=\min_{T}\left\{\sum_{\alpha \in T, \alpha \neq \lambda}H^{T}(G;\alpha)\right\}\text{,}
\end{equation}
\begin{equation}\label{kd_se_node}
    H^{T}(G;\alpha)=-\frac{g_{\alpha}}{vol(G)} \log _{2} \frac{\mathcal{V}_{\alpha}}{\mathcal{V}_{\alpha^{-}}}\text{,}
\end{equation}
where $g_{\alpha}$ is the sum of weights of all edges connecting vertices in $T_{\alpha}$ with vertices outside $T_{\alpha}$.
$\mathcal{V}_{\alpha}$ is the volume of $T_{\alpha}$, the sum of degrees of all vertices in $T_{\alpha}$, and $T$ ranges over all encoding trees of height at most $K$.
The dimension $K$ is also the maximal height of the encoding tree $T$.

\section{The SIRD Optimizing MARL Framework}
In this section, we present the overall framework of the SR-MARL and describe the detailed design of the structural information principles-based role discovery method SIRD.

\begin{figure}[h]
    \centering
    \includegraphics[width=1\columnwidth]{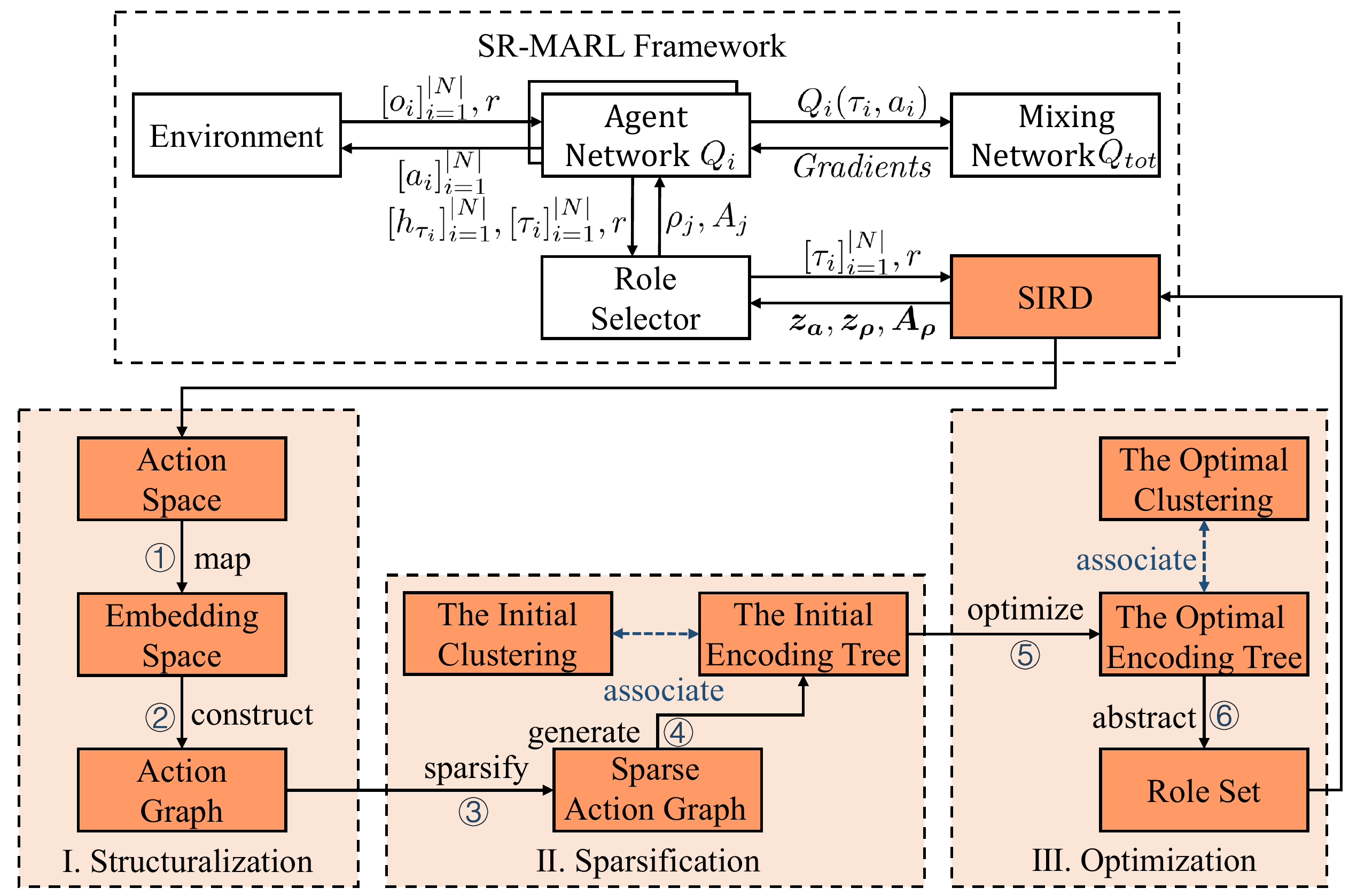}
    \caption{The overall framework of the SR-MARL.}
    \label{SR-MARL}
\end{figure}

\subsection{Overall Framework of SR-MARL}
The SR-MARL consists of five modules: Environment, Agent Network $Q_{i}$, Mixing Network $Q_{tot}$, Role Selector, and the role discovery module SIRD, as shown in Fig.\ref{SR-MARL}.

In the overall framework, each agent $n_{i}$ makes decisions based on individual network $Q_{i}$ which takes the partial observation $o_{i}$ and joint reward $r$ as inputs and is updated by the QPLEX-style mixing network $Q_{tot}$ \cite{wang2020qplex}.
The mixing network $O_{tot}$ has access to global information for centralized training.
Apart from making decisions, the individual network $Q_{i}$ encodes the local action-observation history $\tau_{i}$ into a $d$-dimensional hidden vector $h_{\tau_{i}} \in \mathbb{R}^{d}$, which is then fed into the role selector.
The role discovery module SIRD described in the next subsection takes the action-observation histories of all agents $\left[\tau_{i}\right]_{i=1}^{|N|}$ and joint reward $r$ as inputs.
And the SIRD outputs action representations $\boldsymbol{z_{a}}$, role representations $\boldsymbol{z_{\rho}}$, and restricted action spaces $\boldsymbol{A_{\rho}}$ to the role selector for learning role policies.
Inspired by the RODE \cite{wang2020rode}, the role selector assigns role $\rho_{j} \in \Psi$ and its associated action subspace $A_{j} \in \boldsymbol{A_{\rho}}$ to agent $n_{i}$, based on the dot product between the role representations $\boldsymbol{z_{\rho}}$ and the hidden vector $h_{\tau_{i}}$.
For coordinating the role assignment of all agents, we additionally apply a duplex dueling network of QPLEX \cite{wang2020qplex} to mix the dot products.

\subsection{Role Discovery Module SIRD}
As shown in Fig.\ref{SR-MARL}, the SIRD comprises Structuralization, Sparsification, and Optimization. 
In the structuralization, we map the action space to a fixed-dimensional embedding space and construct an action graph.
In the sparsification, we sparsify the action graph and generate an initial encoding tree of the sparse graph.
In the optimization, we optimize the encoding tree to realize the optimal hierarchical clustering, equivalent to a hierarchical abstracting of actions, and achieve role discovery on the optimal encoding tree.

\begin{figure*}[t]
    \centering
    \includegraphics[width=1\textwidth]{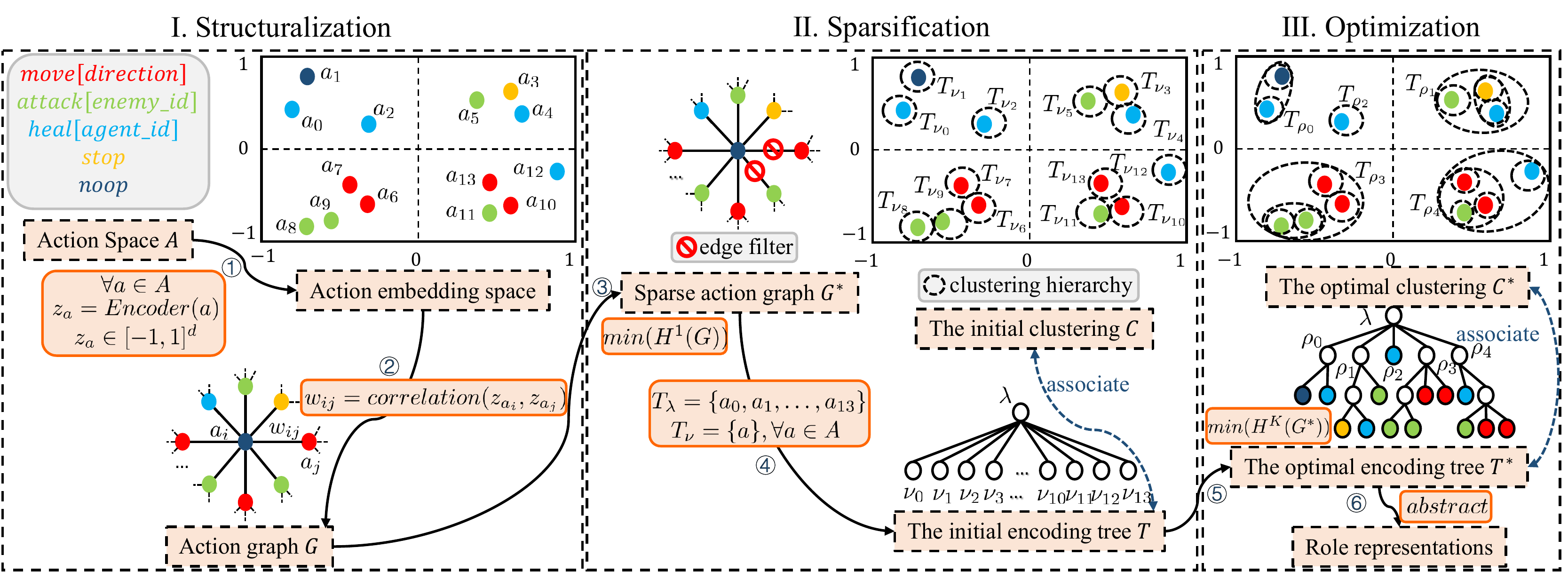}
    \caption{The structural information principles-based role discovery.}
    \label{SERD}
\end{figure*}

\subsubsection{Structuralization.}
Unlike the existing role-based methods \cite{sun2020reinforcement, wang2020roma, wang2020rode}, the SIRD utilizes action correlations to construct an action graph, to improve the effectiveness of the role discovery.
To this end, we learn action representations for achieving team targets, measure correlations between representations, and then construct the weighted, undirected, complete action graph based on the correlations.

Firstly, inspired by the RODE \cite{wang2020rode}, we similarly adopt the encoder-decoder structure \cite{cho2014learning} to learn the action representations, mapping the action space $A$ to a $d$-dimensional embedding space.
In the encoder, we encode each action $a \in A$ as an embedded representation $z_{a} \in \mathbb{R}^{d}$, as the step 1 in Fig.\ref{SERD}\footnote{For better understanding, we set $A=\{a_{0},\dots,a_{13}\}$ and differentiate actions by colors in the action embedding space as an example.}.
For each agent $n_{i}$, the decoder decodes the representation $z_{a_{i}}$ of its adopted action $a_{i}$ to reconstruct its next partial observation $o_{i}^{\prime}$ and joint reward $r$.
Given the action-observation histories of all agents $\left[\tau_{i}\right]_{i=1}^{|N|}$ and joint reward $r$, the encoder-decoder structure is trained end-to-end by minimizing the reconstruction loss.
Secondly, for every two actions $a_{i}$ and $a_{j}$ with $a_{i} \neq a_{j}$, we measure their correlation $\mathcal{C}_{a_{i}, a_{j}}\in\left[-1, 1\right]$ through the Pearson Correlation Analysis:
\begin{equation}\label{pcc}
    \mathcal{C}_{a_{i}, a_{j}}=\frac{E\left((z_{a_{i}}-\mu_{z_{a_{i}}})(z_{a_{j}}-\mu_{z_{a_{j}}})\right)}{\sigma_{z_{a_{i}}} \sigma_{z_{a_{j}}}}\text{,}
\end{equation}
where $\mu_{z_{a_{i}}}$ and $\sigma_{z_{a_{i}}}$ are the mean value and variance of the $d$-dimensional action representation $z_{a_{i}}$, respectively.
Intuitively, the larger absolute value of $\mathcal{C}_{a_{i}, a_{j}}$ represents a more functional correlation between action $a_{i}$ and action $a_{j}$ on achieving team targets.
Thirdly, we take each action in $A$ as a vertex, connect any two vertices $a_{i}$ and $a_{j}$, and assign $\mathcal{C}_{a_{i}, a_{j}}$ to edge $(a_{i},a_{j})$ as weight $w_{ij}$, $w_{ij}=\mathcal{C}_{a_{i}, a_{j}}$, to construct the action graph $G$, as the step 2 in Fig.\ref{SERD}.

Therefore, in the action graph $G$, vertices represent actions in the action space $A$, $V=A$, edges represent action correlations, and edge weight further quantifies the functional correlation between actions.

\subsubsection{Sparsification.}
We implement sparsification of the action graph to reduce the computational cost and eliminate negative interference of trivial weights whose absolute values approach $0$.
Following the construction of cancer cell neighbor networks \cite{li2016three}, we sparsify the action graph $G$ into a $k$-nearest neighbor ($k$-NN) graph $G_{k}$ based on the one-dimensional structural entropy minimization principle.
By minimizing the one-dimensional structural entropy of $G_{k}$, $H^{1}(G_{k})$, we select the optimal number of neighbors $k^{*}$, as the step 3 in Fig.\ref{SERD}.
We summarize the sparsification step as Alg.\ref{alg:sparsification}.

Firstly, based on the mean value $\mu_{w}$ of edge weights in $G$, we introduce a factor $\delta$, $\delta=\frac{1}{2|A|} \cdot \mu_{w}$, $|A|$ is the number of actions, to correct all weights in $G$ (Lines 2-4 in Alg.\ref{alg:sparsification}):
\begin{equation}\label{correction}
    w_{ij}=w_{ij}+\delta\text{.}
\end{equation}
For significant weights whose absolute values approach $1$, the factor $\delta$ is relatively tiny, and the corrected weights are approximately equal to the original values.
For trivial weights, $\delta$ can notably correct them to eliminate their negative interference.
Secondly, for each $k$, we construct the graph $G_{k}$ (Lines 5 and 8 in Alg.\ref{alg:sparsification}) by solely retaining the most significant $k$ edge weights for each vertex and then calculate the one-dimensional structural entropy $H^{1}(G_{k})$ (Lines 6 and 9 in Alg.\ref{alg:sparsification}).
Thirdly, we find all local minimal structural entropies (Lines 10-11 in Alg.\ref{alg:sparsification}), namely \textit{LMSE}, select the minimum $k^{*}$ from the \textit{LMSE}, and output $G_{k^{*}}$ as the sparse action graph $G^{*}$ (Lines 12-14 in Alg.\ref{alg:sparsification}).

\begin{algorithm}[tb]\setcounter{AlgoLine}{0}
    \caption{The Sparsification Algorithm}
    \label{alg:sparsification}
    \LinesNumbered
    \KwIn{The action graph $G$}
    \KwOut{The sparse action graph $G^{*}$}
    \begin{small}
        $\delta\gets$ initialize correction factor
        
        \For{$i=1, \ldots, |A|$}{
        \For{$j=i+1, \ldots, |A|$}{
        $w_{ij}\gets$ correct $w_{ij}$ via Eq.(\ref{correction})
        }
        }
        $G_{1}, G_{2}\gets$ construct $1$-NN graph and $2$-NN graph
        
        $H^{1}(G_{1}), H^{1}(G_{2})\gets$ calculate structural entropy via Eq.(\ref{1d_se})
        
        \For{$k=2, \ldots, |A|-1$}{
        $G_{k+1}\gets$ construct $(k+1)$-NN graph
        
        $H^{1}(G_{k+1}) \gets$ calculate structural entropy of $G_{k+1}$ via Eq.(\ref{1d_se})
        
        \If{$H^{1}(G_{k}) < H^{1}(G_{k-1}) \And H^{1}(G_{k}) < H^{1}(G_{k+1})$}{
        \textit{LMSE} $\gets H^{1}(G_{k})$
        }
        }
        $k^{*} \gets$ find minimial $k$ from \textit{LMSE} 
        
        $G^{*}\gets G_{k^{*}}$
        
        return $G^{*}$
    \end{small}
\end{algorithm}

Therefore, we generate an initial encoding tree $T$ of $G^{*}$: 1) For the action space $A$, we generate a root node $\lambda$ and set its corresponding vertex subset $T_{\lambda}=A$; 2) For each action $a \in A$, we generate a leaf node $\nu$ with $T_{\nu}=\{a\}$ and set $\nu$ as a child node of $\lambda$, $\nu^{-}=\lambda$, as the step 4 in Fig.\ref{SERD}.
Intuitively, the height of the initial encoding tree $T$ is 1.
The associated hierarchical clustering is initialized as a single-level hierarchy where each action is divided into a separate category, as the initial clustering $C$ in Fig.\ref{SERD}.

\subsubsection{Optimization.}
To realize the optimal hierarchical clustering $C^{*}$ of the action space $A$, we optimize the encoding tree $T$ of the sparse action graph $G^{*}$ from $1$ layer to $K$ layers.
Firstly, we introduce two operators \textit{merge} and \textit{combine} from the deDoc \cite{li2018decoding} to minimize the $K$-dimensional structural entropy of the sparse graph $G^{*}$ for optimizing $T$, as the step 5 in Fig.\ref{SERD}.
In the encoding tree $T$, two nodes are brothers if they have a common father node.
The \textit{merge} and \textit{combine} operators are defined on brother nodes and marked as $T_{mg}$ and $T_{cb}$ in our work.
Secondly, based on the $K$-dimensional structural entropy minimization principle \cite{li2018decoding}, we utilize $T_{mg}$ and $T_{cb}$ to design an iterative optimization algorithm, Alg.\ref{alg:optimization}.
At each iteration, Alg.\ref{alg:optimization} traverses all brother nodes $\beta_{1}$ and $\beta_{2}$ in $T$ (Lines 4 and 9 in Alg.\ref{alg:optimization}) and greedily selects operator $T_{mg}$ or $T_{cb}$ that reduces the structural entropy the most (Lines 5 and 10 in Alg.\ref{alg:optimization}) to execute (Lines 7 and 12 in Alg.\ref{alg:optimization}) on the premise that the tree height does not exceed $K$.
When no brother nodes satisfy $\Delta SE > 0$, we terminate the iteration and output $T$ as the optimal encoding tree $T^{*}$.
In the encoding tree $T^{*}$, the root node $\lambda$ corresponds to the action space $A$, $T_{\lambda}=A$, each leaf node $\nu$ corresponds to a singleton containing a single action $a \in A$, and other nodes correspond to clusters of different hierarchies.
For each leaf node $\nu$ with $T_{\nu}=\{a\}$, we set its node representation $z_{\nu}=z_{a}$.
Thirdly, we realize the optimal hierarchical action space clustering $C^{*}$ according to $T^{*}$.

Furthermore, we take the hierarchical clustering $C^{*}$ on the optimal encoding tree $T^{*}$ as hierarchical abstracting of actions to discover roles, as the step 6 in Fig.\ref{SERD}.

\subsubsection{Role Discovery on the Optimal Encoding Tree.}
\begin{figure}[]
    \centering
    \includegraphics[width=1\columnwidth]{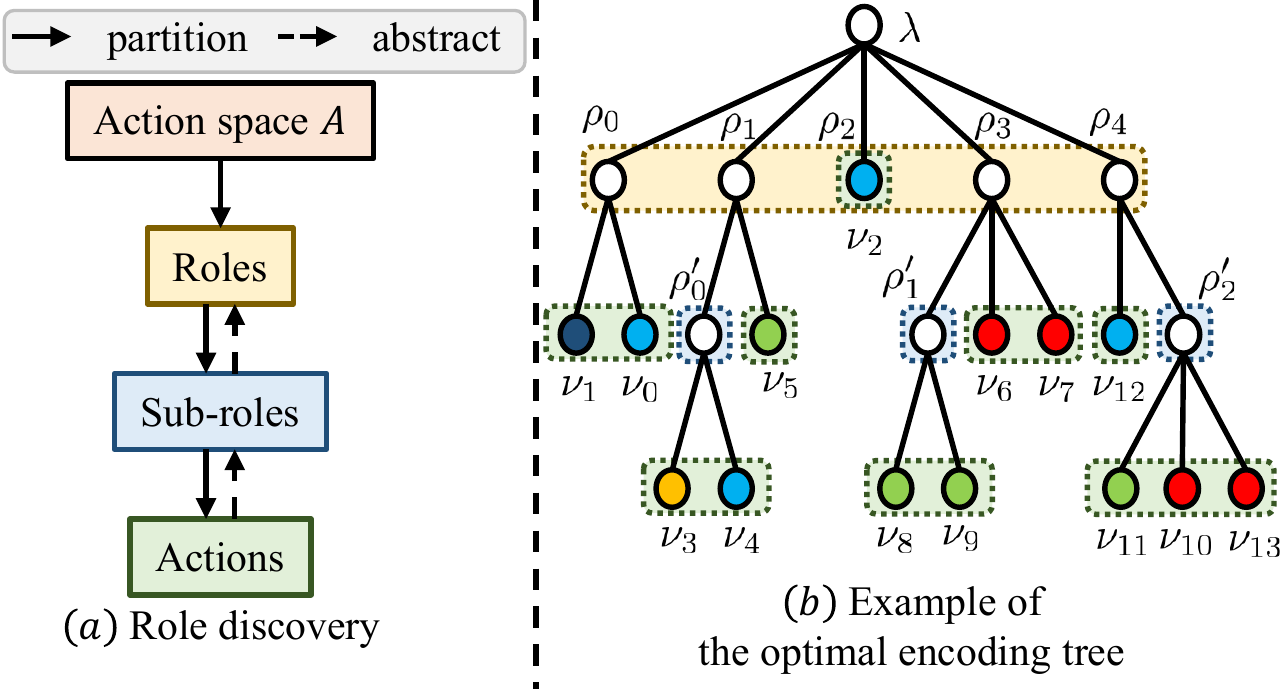}
    \caption{The role discovery on the optimal encoding tree.}
    \label{fig:hierarchical abstraction}
\end{figure}

The optimal encoding tree $T^{*}$ partitions the action space $A$ into a $3$-level abstracting hierarchy from up to bottom, including roles, sub-roles, and actions, as shown in Fig.\ref{fig:hierarchical abstraction}$(a)$.
In the $3$-level hierarchy, children nodes of the root node are defined as roles $\Psi\equiv[\rho_{i}]_{i}^{|\Psi|}$, $\rho_{i}=\lambda^{\wedge}\langle i\rangle$, leaf nodes $[\nu_{i}]_{i}^{|A|}$ are defined as actions, and other nodes on the paths connecting roles and actions are defined as sub-roles $[\rho_{i}^{\prime}]_{i}$, as shown in Fig.\ref{fig:hierarchical abstraction}$(b)$.
For example, actions $\nu_{10}$, $\nu_{11}$, and $\nu_{13}$ are abstracted as a sub-role $\rho_{2}^{\prime}$, and then both $\rho_{2}^{\prime}$ and action $\nu_{12}$ are abstracted as a role $\rho_{4}$.
To calculate role representation $z_{\rho_{i}}$, we need to hierarchically aggregate node representations from bottom to up in the subtree rooted by $\rho_{i}$.
For example, we aggregate action representations $z_{\nu_{10}}$, $z_{\nu_{11}}$, and $z_{\nu_{13}}$ as a sub-role representation $z_{\rho_{2}^{\prime}}$, and then aggregate $z_{\rho_{2}^{\prime}}$ and $z_{\nu_{12}}$ as a role representation $z_{\rho_{4}}$.
To realize aggregating from bottom to up, we propose using the structural entropy distribution as each node's weight.
For each non-leaf node $\alpha \in T^{*}$, the aggregate function is defined to calculate its node representation $z_{\alpha}$ as follows:
\begin{equation}
    z_{\alpha}=\sum_{i=1}^{L} \frac{H^{T^{*}}\left(G ; \alpha^{\wedge}\langle i\rangle\right)}{\sum_{j=1}^{L}{H^{T^{*}}\left(G ; \alpha^{\wedge}\langle j\rangle\right)}} \cdot z_{\alpha^{\wedge}\langle i\rangle}\text{,}
\end{equation}
where $L$ is the number of children nodes of $\alpha$.
In addition, the restricted action space of role $\rho_{i}$ is defined as its corresponding vertex subset $T_{\rho_{i}}$, $A_{i}=T_{\rho_{i}}$.

Finally, we output action representations $\boldsymbol{z_{a}}\equiv\left\{z_{a} \mid \forall a \in A\right\}$, role representations $\boldsymbol{z_{\rho}}\equiv\left\{z_{\rho_{i}} \mid \forall \rho_{i} \in \Psi\right\}$, and restricted action spaces $\boldsymbol{A_{\rho}}\equiv\left\{A_{i} \mid \forall \rho_{i} \in \Psi\right\}$ to achieve the role discovery.

\begin{algorithm}[tb]\setcounter{AlgoLine}{0}
    \caption{The Iterative Optimization Algorithm}
    \label{alg:optimization}
    \KwIn{The initial encoding tree $T$}
    \KwOut{The optimal encoding tree $T^{*}$}
    \begin{small}
        $\Delta SE, \beta_{1}^{*}, \beta_{2}^{*}\gets$ initialization
        
        \While{True}{
        $\Delta SE\gets 0$
        
        \For{each brother nodes $\beta_{1}$ and $\beta_{2}$ in $T$}{
        $\Delta SE, \beta_{1}^{*}, \beta_{2}^{*}\gets$ maximize the reduction of the structural entropy caused by the \textit{merge} operator via Eq.(\ref{kd_se})
        }
        \If{$\Delta SE > 0$}{
        $T\gets T_{mg}(T;\beta_{1}^{*},\beta_{2}^{*})$
        
        Continue
        }
        \For{each brother nodes $\beta_{1}$ and $\beta_{2}$ in $T$}{
        $\Delta SE, \beta_{1}^{*}, \beta_{2}^{*}\gets$ maximize the reduction of the structural entropy caused by the \textit{combine} operator via Eq.(\ref{kd_se})
        }
        \eIf{$\Delta SE > 0$}{
        $T\gets T_{cb}(T;\beta_{1}^{*},\beta_{2}^{*})$
        }
        {
        Break
        }}
        $T^{*}\gets T$
        
        return $T^{*}$
    \end{small}
\end{algorithm}

\subsubsection{Time Complexity of SIRD.}
The overall time complexity of the SIRD is $O\left(n^{2}+n+n \cdot \log ^{2} n\right)$, where $n$ is the number of actions, $n=|A|$.
Here, the time complexity of the structuralization is $O(n^{2})$, which analyzes Pearson Correlation for every two actions.
In the sparsification, the SIRD costs $O(n)$ time complexity to find the minimum $k^{*} \in \{1, \ldots, n\}$ from all local minimal structural entropies.
In our work, the optimization of the encoding tree costs $O(n \cdot \log ^{2} n)$ time complexity by adopting a similar data structure \cite{clauset2004finding}.
Compared with flat clustering in the RODE \cite{wang2020rode}, the SIRD achieves role discovery in the same quadric time complexity and guarantees independency of manual assistance, which leads to effective and stable performance advantages.

\section{Experiments and Analysis}\label{Experiments}
In this section, we conduct a large set of empirical and comparative experiments, aiming to verify the effectiveness and stability of the SR-MARL.
Towards fair evaluation, all results are illustrated with the average and deviation of the performance testing with different random seeds, like in other works \cite{wang2020qplex, wang2020rode}.

\subsection{Experiment Setup}
\subsubsection{Datasets.}
We evaluate the SR-MARL on the StarCraft II micromanagement (SMAC) benchmark \cite{samvelyan2019starcraft}, a mainstream benchmark of CTDE algorithms, of its rich environment and high control complexity.
The SMAC benchmark includes five easy maps, four hard maps, and four super hard maps, where hard and super hard maps are typically hard-exploration tasks requiring agents to learn complex collaboration.
As the SR-MARL is presented for improving multi-agent collaboration, we primarily focus on its performance on hard and super hard maps.
In the micromanagement scenarios, each unit is controlled by an independent agent acting based on local observation, and a built-in AI controls all enemy units.
At each timestep, each agent selects an action from the discrete action space, which consists of moving in four directions, stopping, taking no-op, and selecting an enemy/ally unit to attack/heal.

\subsubsection{Baselines and Variants.}
To make the empirical results more convincing, we compare the SR-MARL, whose maximal encoding tree height is set as 2, with state-of-the-art MARL algorithms, including independent Q-learning method (IQL \cite{tampuu2017multiagent}), value-based methods (VDN \cite{sunehag2017value}, QMIX \cite{rashid2018qmix}, QPLEX \cite{wang2020qplex}, QTRAN \cite{son2019qtran}), actor-critic method (COMA \cite{foerster2018counterfactual}), and role-based method (RODE \cite{wang2020rode}).
The implementations of the SR-MARL and baselines in our experiments are based on the PyMARL (\cite{samvelyan2019starcraft}), and the hyperparameters of the baselines have been fine-tuned on the SMAC benchmark.
All experiments adopt the default settings and are conducted on 3.00GHz Intel Core i9 CPU and NVIDIA RTX A6000 GPU.
\begin{table}[t]
\centering
\resizebox{1\columnwidth}{!}{
\begin{tabular}{|c|c|c|c|c|c|c|c|}
    \hline
    Categories & Easy & Hard & Super Hard \\
    \hline
    COMA & $16.67\pm22.73$ & $4.51$ $\pm$ $\underline{9.58}$ & - \\
    \hline
    IQL & $52.50\pm40.69$ & $73.44\pm24.85$ & $10.55\pm18.49$ \\
    \hline
    VDN & $85.01\pm17.22$ & $71.49\pm18.78$ & $71.10\pm27.23$ \\
    \hline
    QMIX & $\underline{98.44}$ $\pm$ $\underline{2.10}$ & $87.11\pm18.58$ & $70.31\pm38.65$\\
    \hline
    QTRAN & $64.69\pm36.79$ & $58.20\pm45.37$ & $16.80\pm20.61$ \\
    \hline
    QPLEX & $96.88\pm5.04$ & $\underline{89.85}$ $\pm$ $11.35$ & $84.77\pm10.76$\\
    \hline
    RODE & $93.47\pm10.19$ & $88.44\pm20.96$ & $\underline{92.71}$ $\pm$ $\underline{9.20}$ \\
    \hline
    SR-MARL & $\textbf{98.61}\pm\textbf{1.75}$ & $\textbf{95.31}\pm\textbf{6.63}$ & $\textbf{95.71}\pm\textbf{3.10}$\\
    \hline
    \multicolumn{4}{|c|}{Improvements(\%)/Reductions(\%)}\\
    \hline
    Average & $\uparrow0.17$ & $\uparrow6.08$ & $\uparrow3.24$\\
    \hline
    Deviation & $\downarrow16.67$ & $\downarrow30.80$ & $\downarrow66.30$\\
    \hline
\end{tabular}}
\caption{Summary of the test win rates under different map categories: ``average value $\pm$ standard deviation" and ``improvements/reductions" ($\%$). Bold: the best performance under each category, underline: the second performance.}
\label{table:twr}
\end{table}

\subsection{Evaluations}
We evaluate the SR-MARL and state-of-the-art MARL algorithms under different map categories (easy, hard, and super hard) and summarize averages and deviations of test win rates in Table \ref{table:twr}.
Table \ref{table:twr} shows that under different categories, our framework achieves at most a 6.08\% improvement in average value and at most a 66.30\% reduction in deviation.
The improvement in average value and reduction in deviation correspond to the performance advantages on effectiveness and stability of the SR-MARL, respectively.
In terms of effectiveness, explorations in restricted action spaces specified by the role set discovered from the optimal hierarchical abstracting guarantee strong cooperative ability among agents.
In terms of stability, the SR-MARL leverages the structural information principles to guide automatic role discovery and therefore avoids the selection of sensitive hyperparameters.
The performance advantages on effectiveness and stability of the SR-MARL become significant under hard-exploration scenarios (hard and super hard maps).

\begin{figure}[t]
    \centering
    \includegraphics[width=8.6cm, height=3.5cm]{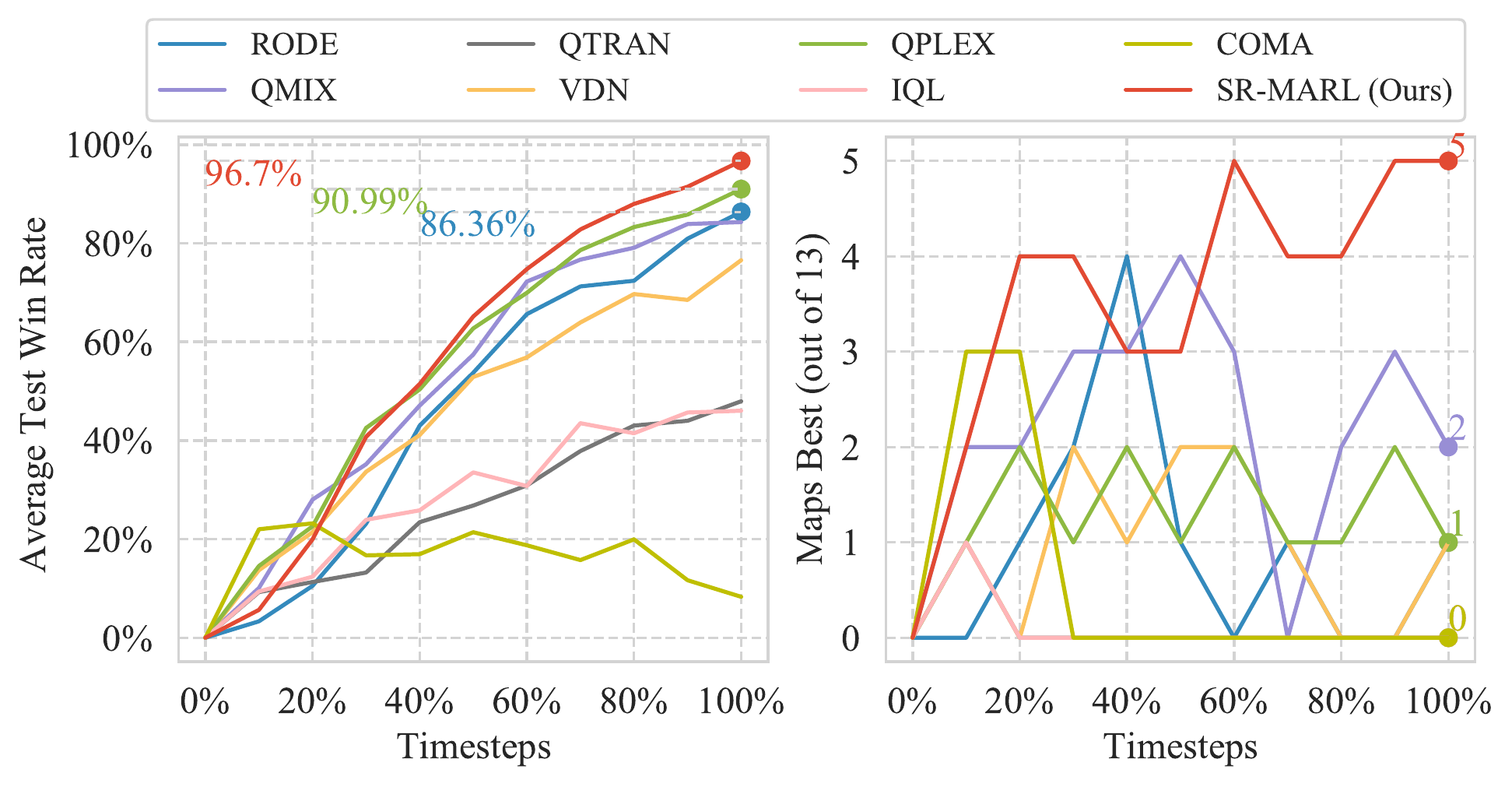}
    \caption{(left) The average test win rates across all 13 maps; (right) the number of maps (out of 13) where the algorithm's average test win rate is the highest.}
    \label{fig:awr_nbm}
\end{figure}

On the other hand, we benchmark the SR-MARL and classical baselines on all 13 maps to evaluate the overall performance across the SMAC suite.
Fig.\ref{fig:awr_nbm} shows each algorithm's average test win rate across all maps and the number of maps where it achieves the highest average test win rate at different stages.
Fig.\ref{fig:awr_nbm}(left) illustrates that compared with baselines, the SR-MARL achieves outstanding overall performance and converges faster.
In particular, the SR-MARL always maintains the highest average test win rate after 40\% of the whole process until obtaining a 96.7\% final average test win rate, which exceeds the second 90.99\% (QPLEX) and the third 86.36\% (RODE) by 5.71\% and 10.34\%.
The advantages of the overall performance and learning efficiency can be attributed to the effective exploration of the action subspaces.
Compared with the RODE, the SR-MARL accomplishes more effective role-based learning via transforming the role discovery into hierarchical action space clustering.  
From Fig.\ref{fig:awr_nbm}(right), the SR-MARL finally performs best on almost half of all maps (5 of 13), much more than baselines.

\begin{figure*}[t]
    \centering
    \includegraphics[width=17cm, height=3.5cm]{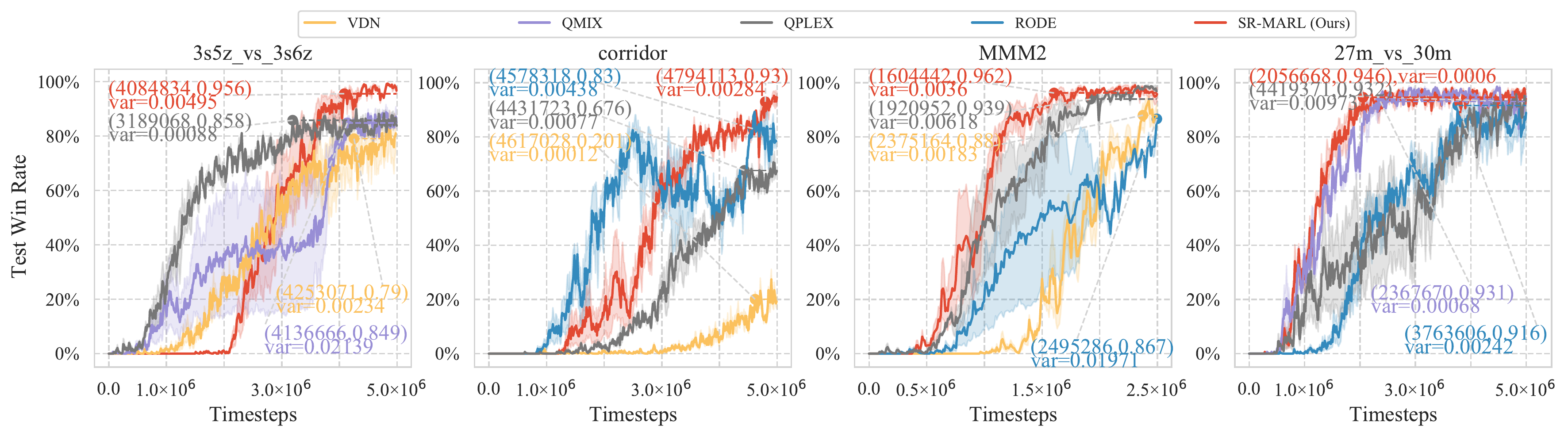}
    \caption{Average test win rates on four SMAC super hard maps.}
    \label{fig:super_hard}
\end{figure*}

In summary, the SR-MARL establishes a new state of the art on SMAC in terms of effectiveness and stability.
Fig.\ref{fig:super_hard} show the learning curves of the SR-MARL and three representative baselines on each super hard map, respectively.
The starting point of convergence and its variance are marked for each curve.
The SR-MARL converges at 2056668 timestep and achieves a 94.6\% average test win rate, with a variance of 0.0006, as shown in $27m\_vs\_30m$.
% \begin{figure*}[t]
%     \centering
%     \includegraphics[width=17cm, height=3.5cm]{hard.pdf}
%     \caption{Average test win rates on four SMAC hard maps.}
%     \label{fig:hard}
% \end{figure*}

\subsubsection{Integrative Abilities.}
\begin{figure}[t]
    \centering
    \includegraphics[width=8.4cm, height=2.7cm]{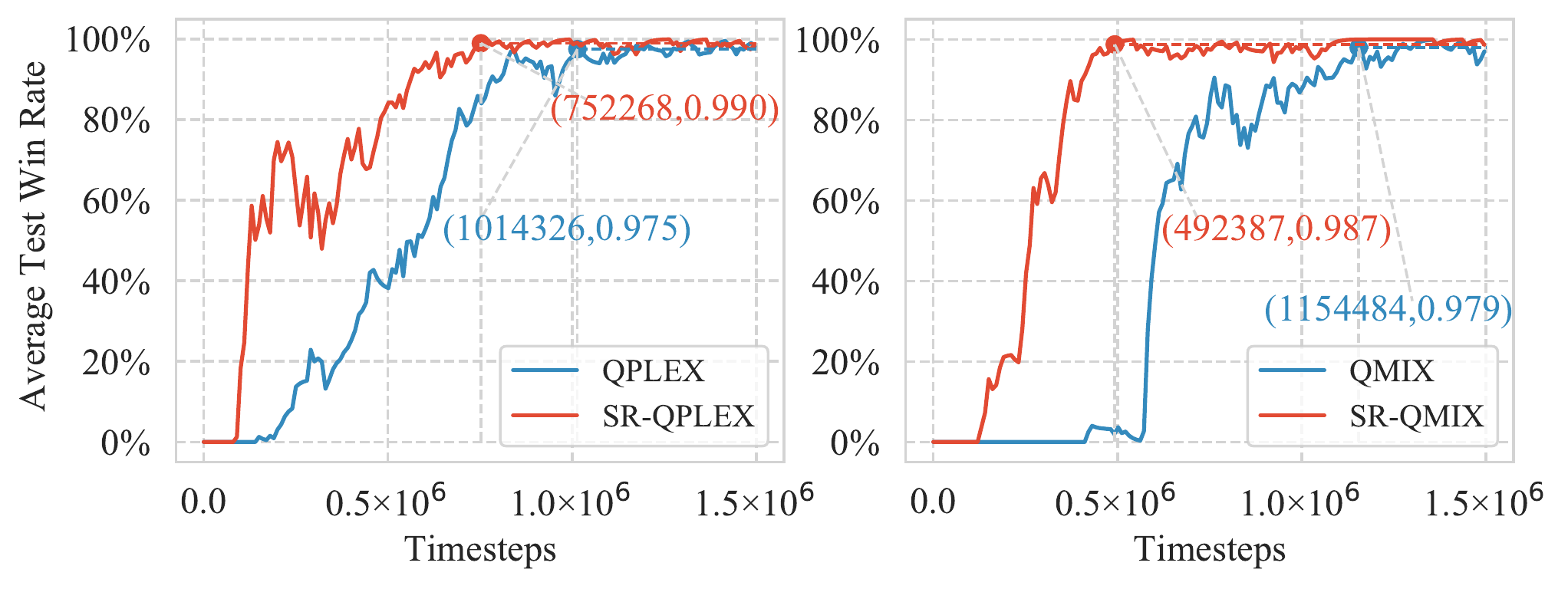}
    \caption{Average test win rates of the SR-MARL integrated with value decomposition methods QMIX and QPLEX.}
    \label{Integrate}
\end{figure}

The SIRD is agnostic to specific MARL algorithms and can be integrated with various value function decomposition approaches by replacing the mixing network $Q_{tot}$ in Fig.\ref{SR-MARL}.
We integrate the SIRD with the QMIX and QPLEX to get SR-QMIX and SR-QPLEX, respectively, and test their performance on map $2c\_vs\_64zg$.
All integrated frameworks outperform the original approaches in effectiveness and efficiency, as shown in Fig.\ref{Integrate}.
The comparative results indicate that the structural information principles-based role discovery method can significantly enhance multi-agent coordination.

\subsubsection{Ablation Studies.}
\begin{figure}[t]
    \centering
    \includegraphics[width=8.4cm, height=2.8cm]{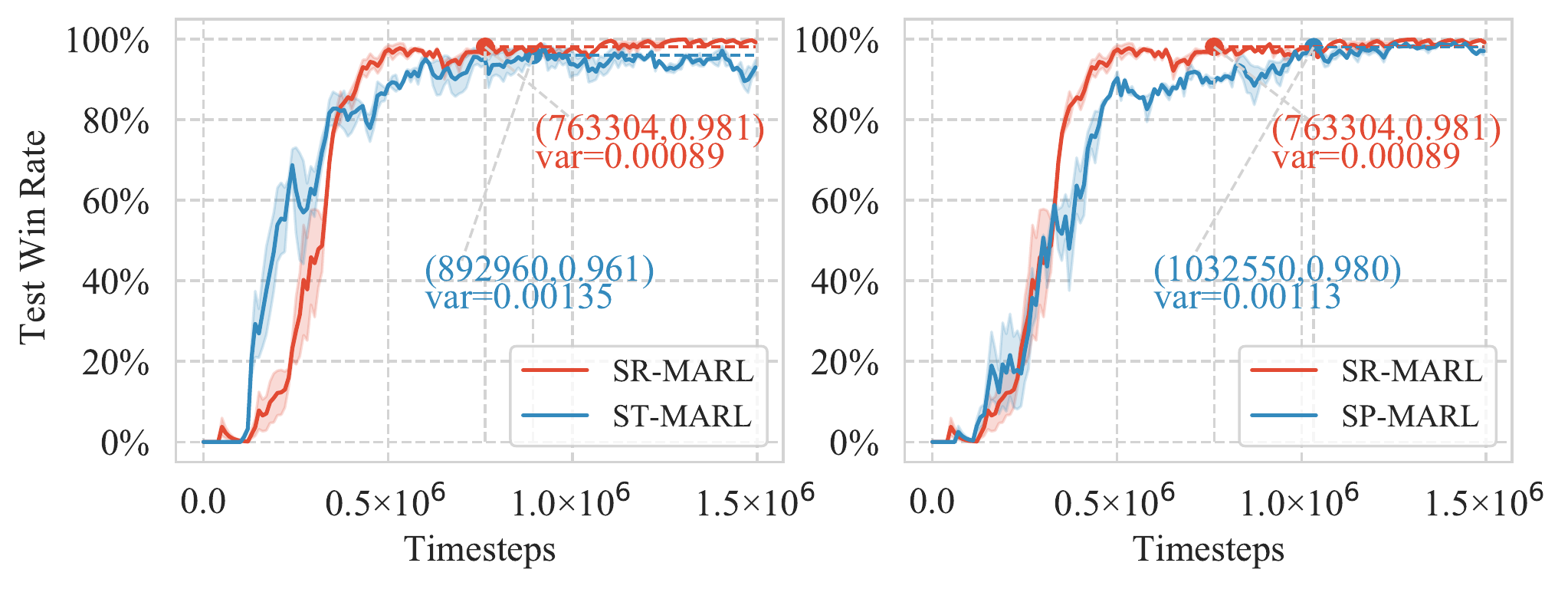}
    \caption{Average test win rates for ablation studies.}
    \label{Ablation}
\end{figure}

We design ablation studies on map $2c\_vs\_64zg$ to evaluate the importance of structuralization and sparsification for performance advantages.
To this end, we design two variants: ST-MARL and SP-MARL, degenerated frameworks of the SR-MARL without some functional modules.
The ST-MARL variant without structuralization discovers roles from the joint action space via K-Means clustering instead of the SIRD module.
The SP-MARL variant without sparsification directly optimizes the encoding tree of the complete action graph for role discovery.
As shown in Fig.\ref{Ablation}, the SR-MARL significantly outperforms the ST-MARL in the average and the deviation of the test win rates, which shows structuralization is the foundation of the SIRD and is crucial for the performance advantages on effectiveness and stability. 
The comparison between SR-MARL and SP-MARL indicates that sparsification remarkably boosts the learning process without affecting the performance advantages.

\begin{figure}[t]
    \centering
    \includegraphics[width=8.4cm, height=2.8cm]{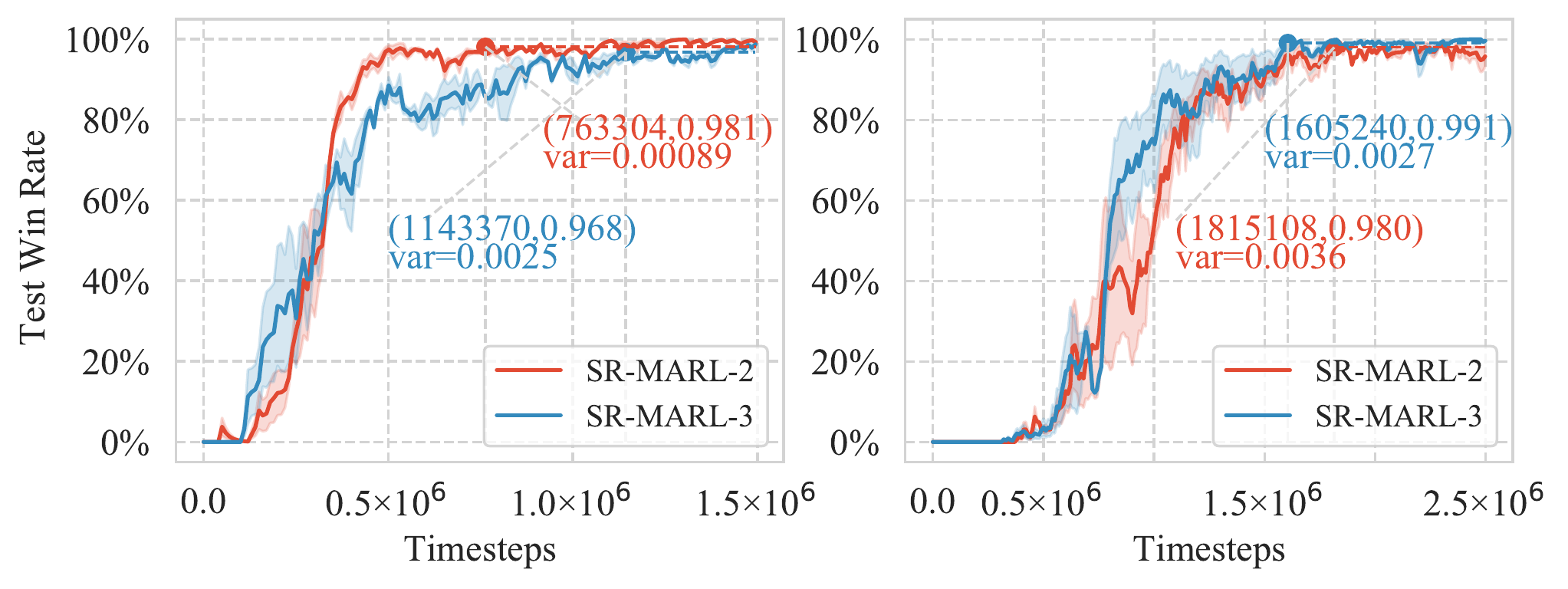}
    \caption{Average test win rates of the SR-MARL with the optimal encoding trees of different maximal heights (2 and 3) on two maps: (left) $2c\_vs\_64zg$ and (right) $MMM2$.}
    \label{Ablation_OP}
\end{figure}

We discuss different maximal tree heights (2 and 3) and separately name associated frameworks SR-MARL-2 and SR-MARL-3.
Fig.\ref{Ablation_OP} shows their learning curves on two different maps, $2c\_vs\_64zg$ and $MMM2$.
In terms of effectiveness, stability, and efficiency, the SR-MARL-3 performs better on super hard map $MMM2$, while the SR-MARL-2 performs better on hard map $2c\_vs\_64zg$.
The reason is maybe that super hard maps require a more hierarchical action space abstracting associated with a higher encoding tree to achieve complex multi-agent collaboration.

\section{Related Work}\label{Related Work}
\subsubsection{Role-based Learning.}
In natural systems \cite{gordon1996organization, jeanson2005emergence, butler2012condensed}, the role is closely related to labor division and efficiency improvement.
Inspired by the above benefits, multi-agent systems decompose the task and specialize agents with the same role to subtasks for reducing the design complexity \cite{wooldridge2000gaia, cossentino2014handbook}.
However, these methodologies use predefined task decomposition and roles that are unavailable in practice \cite{lhaksmana2018role, sun2020reinforcement}.
Bayesian inference is introduced in MARL to learn roles \cite{wilson2010bayesian}, and ROMA encourages the emergence of roles by designing a specialization objective \cite{wang2020roma}.
Searching in the entire state-action space makes these methods inefficient.
RODE achieves efficient role discovery by first decomposing joint action spaces \cite{wang2020rode}.
Instead of flat clustering in RODE, the proposed SIRD utilizes hierarchical action space clustering to achieve more effective and stable role discovery.
% \subsubsection{Structural Information Principles.}
% Structural information principles were first proposed in 2016 \cite{li2016structural}.
% % including definitions of partitioning tree and structural entropy
% % The structural entropy measures the uncertainty of a graph, and the partitioning tree achieves a hierarchical partitioning of the graph.
% Li et al. \cite{li2016three} minimize the one-dimensional structural entropy to construct a cancer cell neighbor network.
% % Based on the $K$-dimensional structural entropy minimization principle \cite{li2018decoding}, deDoc decodes topologically associating domains with ultra-low resolution Hi-C data.
% By minimizing the $K$-dimensional structural entropy \cite{li2018decoding}, deDoc decodes topologically associating domains with ultra-low resolution Hi-C data.
% And the dynamics measurement of the encoding tree is recently analyzed \cite{Yang2022DynamicMO}. 

% \input{7-conclusion}
\section{Conclusion}\label{Conclusion}
This paper proposes a structural information principles-based role discovery method SIRD and presents a SIRD optimizing MARL framework SR-MARL for multi-agent collaboration.
To the best of our knowledge, this is the first time that the mathematical structural information principles have been incorporated into MARL to improve performance under cooperative scenarios.
Evaluations of challenging tasks in the SMAC benchmark demonstrate that the SR-MARL shows significant outperformance on effectiveness and stability over the state-of-the-art baselines and even can be flexibly integrated with various value function decomposition approaches to enhance coordination.
For future work, we will conduct more analysis and explorations on the encoding tree under MARL scenarios.

\section*{Acknowledgments}
The corresponding authors are Hao Peng and Angsheng Li. 
This paper was supported by the National Key R\&D Program of China through grant 2021YFB1714800, NSFC through grant 61932002, S\&T Program of Hebei through grant 20310101D, Natural Science Foundation of Beijing Municipality through grant 4222030, and the Fundamental Research Funds for the Central Universities. 
\bibliography{0-main}

\begin{thebibliography}{41}
\providecommand{\natexlab}[1]{#1}

\bibitem[{Baker et~al.(2020)Baker, Kanitscheider, Markov, Wu, Powell, McGrew,
  and Mordatch}]{baker2019emergent}
Baker, B.; Kanitscheider, I.; Markov, T.~M.; Wu, Y.; Powell, G.; McGrew, B.;
  and Mordatch, I. 2020.
\newblock Emergent Tool Use From Multi-Agent Autocurricula.
\newblock In \emph{Proceedings of the International Conference on Learning
  Representations}, 1--28.

\bibitem[{Bonjean et~al.(2014)Bonjean, Mefteh, Gleizes, Maurel, and
  Migeon}]{cossentino2014handbook}
Bonjean, N.; Mefteh, W.; Gleizes, M.; Maurel, C.; and Migeon, F. 2014.
\newblock \emph{Handbook on Agent-oriented Design Processes}.
\newblock Springer.

\bibitem[{Butler(2012)}]{butler2012condensed}
Butler, E. 2012.
\newblock \emph{The Condensed Wealth of Nations}.
\newblock Centre for Independent Studies.

\bibitem[{Cho et~al.(2014)Cho, Merri{\"e}nboer, Gulcehre, Bahdanau, Bougares,
  Schwenk, and Bengio}]{cho2014learning}
Cho, K.; Merri{\"e}nboer, B.~V.; Gulcehre, C.; Bahdanau, D.; Bougares, F.;
  Schwenk, H.; and Bengio, Y. 2014.
\newblock Learning phrase representations using RNN encoder-decoder for
  statistical machine translation.
\newblock In \emph{Proceedings of the 2014 Conference on Empirical Methods in
  Natural Language Processing}, 1724--1734.

\bibitem[{Claus and Boutilier(1998)}]{claus1998dynamics}
Claus, C.; and Boutilier, C. 1998.
\newblock The Dynamics of Reinforcement Learning in Cooperative Multiagent
  Systems.
\newblock In \emph{Proceedings of the Fifteenth National Conference on
  Artificial Intelligence and Tenth Innovative Applications of Artificial
  Intelligence Conference}, 746--752.

\bibitem[{Clauset, Newman, and Moore(2004)}]{clauset2004finding}
Clauset, A.; Newman, M.~E.; and Moore, C. 2004.
\newblock Finding community structure in very large networks.
\newblock \emph{Physical Review E}, 70: 066111.

\bibitem[{DeLoach and Garc{\'{\i}}a{-}Ojeda(2010)}]{deloach2010mase}
DeLoach, S.~A.; and Garc{\'{\i}}a{-}Ojeda, J.~C. 2010.
\newblock O-MaSE: a customisable approach to designing and building complex,
  adaptive multi-agent systems.
\newblock \emph{International Journal of Agent-Oriented Software Engineering},
  4(3): 244--280.

\bibitem[{Foerster et~al.(2018)Foerster, Farquhar, Afouras, Nardelli, and
  Whiteson}]{foerster2018counterfactual}
Foerster, J.~N.; Farquhar, G.; Afouras, T.; Nardelli, N.; and Whiteson, S.
  2018.
\newblock Counterfactual Multi-Agent Policy Gradients.
\newblock In \emph{Proceedings of the AAAI Conference on Artificial
  Intelligence}, 2974--2982.

\bibitem[{Gordon(1996)}]{gordon1996organization}
Gordon, D.~M. 1996.
\newblock The organization of work in social insect colonies.
\newblock \emph{Nature}, 380(6570): 121--124.

\bibitem[{Gupta, Egorov, and Kochenderfer(2017)}]{gupta2017cooperative}
Gupta, J.~K.; Egorov, M.; and Kochenderfer, M.~J. 2017.
\newblock Cooperative Multi-agent Control Using Deep Reinforcement Learning.
\newblock In \emph{Proceedings of the International Conference on Autonomous
  Agents and Multiagent Systems}, 66--83.

\bibitem[{Jeanson, Kukuk, and Fewell(2005)}]{jeanson2005emergence}
Jeanson, R.; Kukuk, P.~F.; and Fewell, J.~H. 2005.
\newblock Emergence of division of labour in halictine bees: contributions of
  social interactions and behavioural variance.
\newblock \emph{Animal Behaviour}, 70(5): 1183--1193.

\bibitem[{Karami and Johansson(2014)}]{karami2014choosing}
Karami, A.; and Johansson, R. 2014.
\newblock Choosing DBSCAN parameters automatically using differential
  evolution.
\newblock \emph{International Journal of Computer Applications}, 91(7): 1--11.

\bibitem[{Kraemer and Banerjee(2016)}]{kraemer2016multi}
Kraemer, L.; and Banerjee, B. 2016.
\newblock Multi-agent reinforcement learning as a rehearsal for decentralized
  planning.
\newblock \emph{Neurocomputing}, 190: 82--94.

\bibitem[{Laurent et~al.(2011)Laurent, Matignon, Fort-Piat
  et~al.}]{laurent2011world}
Laurent, G.~J.; Matignon, L.; Fort-Piat, L.; et~al. 2011.
\newblock The world of independent learners is not markovian.
\newblock \emph{International Journal of Knowledge-based and Intelligent
  Engineering Systems}, 15: 55--64.

\bibitem[{Lhaksmana, Murakami, and Ishida(2018)}]{lhaksmana2018role}
Lhaksmana, K.~M.; Murakami, Y.; and Ishida, T. 2018.
\newblock Role-based modeling for designing agent behavior in self-organizing
  multi-agent systems.
\newblock \emph{International Journal of Software Engineering and Knowledge
  Engineering}, 28(01): 79--96.

\bibitem[{Li and Pan(2016)}]{li2016structural}
Li, A.; and Pan, Y. 2016.
\newblock Structural Information and Dynamical Complexity of Networks.
\newblock \emph{IEEE Transactions on Information Theory}, 62: 3290--3339.

\bibitem[{Li, Yin, and Pan(2016)}]{li2016three}
Li, A.; Yin, X.; and Pan, Y. 2016.
\newblock Three-dimensional gene map of cancer cell types: Structural entropy
  minimisation principle for defining tumour subtypes.
\newblock \emph{Scientific Reports}, 6: 1--26.

\bibitem[{Li et~al.(2018)Li, Yin, Xu, Wang, Han, Wei, Deng, Xiong, and
  Zhang}]{li2018decoding}
Li, A.; Yin, X.; Xu, B.; Wang, D.; Han, J.; Wei, Y.; Deng, Y.; Xiong, Y.; and
  Zhang, Z. 2018.
\newblock Decoding topologically associating domains with ultra-low resolution
  Hi-C data by graph structural entropy.
\newblock \emph{Nature Communications}, 9: 1--12.

\bibitem[{Mahajan et~al.(2019)Mahajan, Rashid, Samvelyan, and
  Whiteson}]{mahajan2019maven}
Mahajan, A.; Rashid, T.; Samvelyan, M.; and Whiteson, S. 2019.
\newblock Maven: Multi-agent variational exploration.
\newblock \emph{Advances in Neural Information Processing Systems}, 32:
  7611--7622.

\bibitem[{Nguyen, Nguyen, and Nahavandi(2020)}]{nguyen2020deep}
Nguyen, T.~T.; Nguyen, N.~D.; and Nahavandi, S. 2020.
\newblock Deep reinforcement learning for multiagent systems: A review of
  challenges, solutions, and applications.
\newblock \emph{IEEE Transactions on Cybernetics}, 50(9): 3826--3839.

\bibitem[{Now{\'e}, Vrancx, and Hauwere(2012)}]{nowe2012game}
Now{\'e}, A.; Vrancx, P.; and Hauwere, Y.-M.~D. 2012.
\newblock Game Theory and Multi-agent Reinforcement Learning.
\newblock In \emph{Reinforcement Learning}, 441--470. Springer.

\bibitem[{Oliehoek and Amato(2016)}]{oliehoek2016concise}
Oliehoek, F.~A.; and Amato, C. 2016.
\newblock \emph{A Concise Introduction to Decentralized POMDPs}.
\newblock Springer.

\bibitem[{Oliehoek, Spaan, and Vlassis(2008)}]{oliehoek2008optimal}
Oliehoek, F.~A.; Spaan, M. T.~J.; and Vlassis, N. 2008.
\newblock Optimal and approximate Q-value functions for decentralized POMDPs.
\newblock \emph{Journal of Artificial Intelligence Research}, 32: 289--353.

\bibitem[{Peng et~al.(2022)Peng, Zhang, Li, Cao, Pan, and
  Yu}]{peng2022reinforced}
Peng, H.; Zhang, R.; Li, S.; Cao, Y.; Pan, S.; and Yu, P. 2022.
\newblock Reinforced, incremental and cross-lingual event detection from social
  messages.
\newblock \emph{IEEE Transactions on Pattern Analysis and Machine
  Intelligence}.

\bibitem[{Rashid et~al.(2018)Rashid, Samvelyan, de~Witt, Farquhar, Foerster,
  and Whiteson}]{rashid2018qmix}
Rashid, T.; Samvelyan, M.; de~Witt, C.~S.; Farquhar, G.; Foerster, J.~N.; and
  Whiteson, S. 2018.
\newblock QMIX: Monotonic Value Function Factorisation for Deep Multi-Agent
  Reinforcement Learning.
\newblock In \emph{Proceedings of the International Conference on Machine
  Learning}, 4292--4301.

\bibitem[{Samvelyan et~al.(2019)Samvelyan, Rashid, de~Witt, Farquhar, Nardelli,
  Rudner, Hung, Torr, Foerster, and Whiteson}]{samvelyan2019starcraft}
Samvelyan, M.; Rashid, T.; de~Witt, C.~S.; Farquhar, G.; Nardelli, N.; Rudner,
  T. G.~J.; Hung, C.; Torr, P. H.~S.; Foerster, J.~N.; and Whiteson, S. 2019.
\newblock The StarCraft Multi-Agent Challenge.
\newblock In \emph{Proceedings of the International Conference on Autonomous
  Agents and Multiagent Systems}, 2186--2188.

\bibitem[{Son et~al.(2019)Son, Kim, Kang, Hostallero, and Yi}]{son2019qtran}
Son, K.; Kim, D.; Kang, W.~J.; Hostallero, D.; and Yi, Y. 2019.
\newblock QTRAN: Learning to Factorize with Transformation for Cooperative
  Multi-Agent Reinforcement Learning.
\newblock In \emph{Proceedings of the International Conference on Machine
  Learning}, 5887--5896.

\bibitem[{Sun, Liu, and Dong(2020)}]{sun2020reinforcement}
Sun, C.; Liu, W.; and Dong, L. 2020.
\newblock Reinforcement learning with task decomposition for cooperative
  multiagent systems.
\newblock \emph{IEEE Transactions on Neural Networks and Learning Systems},
  32(5): 2054--2065.

\bibitem[{Sunehag et~al.(2018)Sunehag, Lever, Gruslys, Czarnecki, Zambaldi,
  Jaderberg, Lanctot, Sonnerat, Leibo, Tuyls, and Graepel}]{sunehag2017value}
Sunehag, P.; Lever, G.; Gruslys, A.; Czarnecki, W.~M.; Zambaldi, V.~F.;
  Jaderberg, M.; Lanctot, M.; Sonnerat, N.; Leibo, J.~Z.; Tuyls, K.; and
  Graepel, T. 2018.
\newblock Value-Decomposition Networks For Cooperative Multi-Agent Learning
  Based On Team Reward.
\newblock In \emph{Proceedings of the International Conference on Autonomous
  Agents and MultiAgent Systems}, 2085--2087.

\bibitem[{Tampuu et~al.(2017)Tampuu, Matiisen, Kodelja, Kuzovkin, Korjus, Aru,
  Aru, and Vicente}]{tampuu2017multiagent}
Tampuu, A.; Matiisen, T.; Kodelja, D.; Kuzovkin, I.; Korjus, K.; Aru, J.; Aru,
  J.; and Vicente, R. 2017.
\newblock Multiagent Cooperation and Competition with Deep Reinforcement
  Learning.
\newblock \emph{PloS One}, 12: e0172395.

\bibitem[{Tan(1993)}]{tan1993multi}
Tan, M. 1993.
\newblock Multi-Agent Reinforcement Learning: Independent versus Cooperative
  Agents.
\newblock In \emph{Proceedings of the International Conference on Machine
  Learning}, 330--337.

\bibitem[{Vinyals et~al.(2019)Vinyals, Babuschkin, Czarnecki, and
  etc.}]{AlphaStar2019}
Vinyals, O.; Babuschkin, I.; Czarnecki, W.~M.; and etc. 2019.
\newblock AlphaStar: Grandmaster level in StarCraft II using multi-agent
  reinforcement learning.
\newblock \emph{Nature}, 575(7782): 350--354.

\bibitem[{Wang et~al.(2021{\natexlab{a}})Wang, Ren, Liu, Yu, and
  Zhang}]{wang2020qplex}
Wang, J.; Ren, Z.; Liu, T.; Yu, Y.; and Zhang, C. 2021{\natexlab{a}}.
\newblock Qplex: Duplex Dueling Multi-Agent Q-Learning.
\newblock In \emph{Proceedings of the International Conference on Learning
  Representations}, 1--27.

\bibitem[{Wang et~al.(2020)Wang, Dong, Lesser, and Zhang}]{wang2020roma}
Wang, T.; Dong, H.; Lesser, V.~R.; and Zhang, C. 2020.
\newblock ROMA: Multi-Agent Reinforcement Learning with Emergent Roles.
\newblock In \emph{Proceedings of the International Conference on Machine
  Learning}, 9876--9886.

\bibitem[{Wang et~al.(2021{\natexlab{b}})Wang, Gupta, Mahajan, Peng, Whiteson,
  and Zhang}]{wang2020rode}
Wang, T.; Gupta, T.; Mahajan, A.; Peng, B.; Whiteson, S.; and Zhang, C.
  2021{\natexlab{b}}.
\newblock Rode: Learning Roles to Decompose Multi-Agent Tasks.
\newblock In \emph{Proceedings of the International Conference on Learning
  Representations}, 1--24.

\bibitem[{Wilson, Fern, and Tadepalli(2010)}]{wilson2010bayesian}
Wilson, A.; Fern, A.; and Tadepalli, P. 2010.
\newblock Bayesian policy search for multi-agent role discovery.
\newblock In \emph{Proceedings of the AAAI Conference on Artificial
  Intelligence}, 624--629.

\bibitem[{Wooldridge, Jennings, and Kinny(2000)}]{wooldridge2000gaia}
Wooldridge, M.~J.; Jennings, N.~R.; and Kinny, D. 2000.
\newblock The Gaia methodology for agent-oriented analysis and design.
\newblock \emph{Autonomous Agents and Multi-Agent Systems}, 3(3): 285--312.

\bibitem[{Yang, Peng, and Li(2022)}]{Yang2022DynamicMO}
Yang, R.; Peng, H.; and Li, A. 2022.
\newblock Dynamic Measurement of Structural Entropy for Dynamic Graphs.
\newblock \emph{ArXiv}, abs/2207.12653.

\bibitem[{Yang and Wang(2020)}]{yang2020overview}
Yang, Y.; and Wang, J. 2020.
\newblock An overview of multi-agent reinforcement learning from game
  theoretical perspective.
\newblock \emph{arXiv preprint arXiv:2011.00583}.

\bibitem[{Zhang and Lesser(2011)}]{zhang2011coordinated}
Zhang, C.; and Lesser, V.~R. 2011.
\newblock Coordinated Multi-Agent Reinforcement Learning in Networked
  Distributed POMDPs.
\newblock In \emph{Proceedings of the AAAI Conference on Artificial
  Intelligence}, 764--770.

\bibitem[{Zhang et~al.(2022)Zhang, Peng, Dou, Wu, Sun, Li, Zhang, and
  Yu}]{zhang2022automating}
Zhang, R.; Peng, H.; Dou, Y.; Wu, J.; Sun, Q.; Li, Y.; Zhang, J.; and Yu, P.~S.
  2022.
\newblock Automating DBSCAN via Deep Reinforcement Learning.
\newblock In \emph{Proceedings of the 31st ACM International Conference on
  Information \& Knowledge Management}, 2620--2630.

\end{thebibliography}
\end{document}